\documentclass[twocolumn,a4paper]{article}
\usepackage{nolta2020}
\usepackage{amsmath} 
\usepackage{txfonts}
\usepackage{algorithm}
\usepackage{graphicx}
\usepackage{algpseudocode}

\begin{document}

\title{A Nesterov's Accelerated quasi-Newton method for Global Routing using Deep Reinforcement Learning}

\author{S. Indrapriyadarsini${}^\dag$ , Shahrzad Mahboubi${}^\S$, Hiroshi Ninomiya${}^\S$, Takeshi Kamio${}^*$,  Hideki Asai${}^\ddag$}

\address{
${}^\dag$Graduate School of Science and Technology /
${}^\ddag$Research Institute of Electronics, Shizuoka University\\
3-5-1 Johoku, Naka-ku, Hamamatsu, Shizuoka Prefecture 432-8011, Japan\\
${}^\S$Graduate School of Electrical and Information Engineering, Shonan Institute of Technology\\
1-1-25 Tsujidonishikaigan, Fujisawa, Kanagawa Prefecture 251-0046, Japan\\
\* ${}^*$ Graduate School of Information Sciences, Hiroshima City University, \\3-4-1 Ozukahigashi, Asaminami Ward, Hiroshima, 731-3194, Japan\\
Email: s.indrapriyadarsini.17@shizuoka.ac.jp, 20T2502@sit.shonan-it.ac.jp, ninomiya@info.shonan-it.ac.jp, \\kamio@hiroshima-cu.ac.jp, asai.hideki@shizuoka.ac.jp
}

\maketitle

\abstract
Deep Q-learning method is one of the most popularly used deep reinforcement learning algorithms which uses deep neural networks to approximate the estimation of the action-value function. Training of the deep Q-network (DQN) is usually restricted to first order gradient based methods. This paper attempts to accelerate the training of deep Q-networks by introducing a second order Nesterov's accelerated quasi-Newton method. We evaluate the performance of the proposed method on deep reinforcement learning using double DQNs for global routing. The results show that the proposed method can obtain better routing solutions compared to the DQNs trained with first order Adam and RMSprop methods.
\endabstract

\vspace{-2mm}
\section{Introduction}
Reinforcement learning (RL) is a machine learning technique where an agent perceives its current state $s$ and takes actions $a$ by interacting with an environment. The environment, in return provides a reward $\mathcal{R}$,  while the reinforcement learning algorithm attempts to find a policy $\pi$ for maximizing the cumulative reward for the agent over the course of the problem \cite{Sutton1998}. The Q-learning algorithm is one of the popular off-policy reinforcement learning algorithms that chooses the best action based on estimates of the state-action value $Q(s,a)$ represented in the form of a table called the Q-table. As the state and action space of the problem increases, the estimation of the state-action value can be slow and time consuming and hence estimated as a function approximation. These function approximations can be represented as a non-convex, non-linear unconstrained optimization problem and can be solved using deep neural networks (known as deep Q-networks). 

Training of Deep Q-Networks (DQN) are usually restricted to first order methods such as stochastic gradient descent (SGD), Adam \cite{kingma2014adam}, RMSprop \cite{RMSprop2012} etc. Using second order curvature information have shown to improve the performance and convergence speed for non convex optimization problems \cite{Nocedal2006}. The BFGS method is one of the most popular second order quasi-Newton method. The Nesterov's accelerated quasi-Newton (NAQ) method \cite{ninomiya2017novel} was shown to accelerate the BFGS method using the Nesterov's accelerated gradient term. In extension to our previous work in \cite{aSNAQ2019}, we show that the adaptive stochastic Nesterov's accelerated quasi-Newton method (aSNAQ) allows more stable approximations and is efficient in training DQNs for deep reinforcement learning applications. We evaluate the performance of the proposed method in comparison to popular first order methods in solving the global routing problem using reinforcement learning.

Synthesis and physical design optimizations are the core tasks of the VLSI / ASIC design flow. Global routing has been a challenging problem in IC physical design. Given a netlist with the description of all the components, their connections and positions, the goal of the router is to determine the path for all the connections without violating the constraints and design rules. Conventional routing automation tools are usually based on analytical and path search algorithms which are NP complete. Hence a machine learning approach would be more suitable for this kind of automation problem.  Studies that propose AI techniques such as machine learning, deep learning, genetic algorithms deal with only prediction of routability, short violations, pin-access violations, etc.  Moreover, the non-availability of large labelled training datasets for a supervised learning model is another challenge. Thus deep reinforcement learning (DRL) is a potential approach to such applications.  Recently, a reinforcement learning approach to global routing that uses first order gradient based method for training the DQN was proposed in \cite{Liao2020}.  

This paper attempts to accelerate training of deep Q-networks by introducing a second order Nesterov's accelerated quasi-Newton method  to get better routing solutions using a deep reinforcement learning approach. Also, to further enhance the performance of the DRL model for global routing, we use double deep Q-learning \cite {DoubleDQN2016}. The obtained routing solution is evaluated in terms of total wirelength and overflow and compared with the results of Adam and RMSprop.

\section {Background}
The Reinforcement Learning problem is modelled as a Markov's Decision Process (MDP). In order to solve the MDP, the estimates of the value function of all possible actions is learnt using Q-learning method, a form of temporal difference learning \cite{Nocedal2006}.  The optimal action-value function $Q^*(s,a)$ that maximizes the cummulative reward satisfies the Bellman equation and is given as\vspace{-1mm}
\begin{equation}  
{Q^*(s,a)} = E_{s' \thicksim \zeta} [ \mathcal{R}+ \gamma ~{\rm max}_{a'}Q^*(s', a') | s,a ]\vspace{-1mm}
\end{equation}
where $\gamma$ is the discount factor. 
The Q-learning algorithm is an off-policy, model-free reinforcement learning algorithm that iteratively learns the optimal action-value function. In deep Q-learning, this function is optimized by a neural network parameterized by ${\bf w}$. The inputs to the neural network are the states $s$ and the output predicted by the neural network correspond to the action values $Q(s,a;{\bf w})$ for each action $a$.  Deep Q-networks use experience replay to train the network. A replay buffer of fixed memory $\mathcal{D}$ stores the transitions $(s,a,r,s')$ of the agent's experiences from which samples are randomly drawn for training the DQN. The loss function $\mathcal{L}({\bf w})$ as shown in (\ref{loss}) is used in backpropagation and calculation of the gradients for updating the parameters ${\bf w}$. \vspace{-1mm}
\begin{equation}\label{loss}
{\mathcal{L}({\bf w})} = E_{(s,a) \thicksim \zeta} [(\mathcal{Y} - Q_{\bf w}(s, a))^2 ]\vspace{-1mm}
\end{equation}
where the target function $\mathcal{Y}_i$ is given as\vspace{-1mm}
\begin{equation}  
\mathcal{Y} = E_{(s') \thicksim \zeta} [\mathcal{R} + \gamma ~{\rm max}_{a'}Q_{{\bf w}}(s', a') ]\vspace{-1mm}
\end{equation}
DQNs are said to overestimate the update of the action-value function since  $Q_{\bf w}(s, a)$ is used to select the best next action at state $s'$ and apply the action value predicted by the same $Q_{\bf w}(s, a)$. Double Deep Q-learning \cite{DoubleDQN2016} resolves this issue by decoupling the action selection and action value estimation using two Q-networks.  \vspace{-1mm}
\begin{equation}  
\mathcal{Y} = E_{(s') \thicksim \zeta} [\mathcal{R} + \gamma Q_{{\bf w}^-}(s', {\rm argmax}_{a'}Q_{{\bf w}}(s', a')) ]\vspace{-1mm}
\end{equation}
${\bf w}$ and ${\bf w}^-$ represent the parameters of the two Q networks -- primary and target networks respectively. The primary network parameters are periodically copied to the target network using Polyak averaging (\ref{poly}), where $\tau$ is a small value such as 0.05.\vspace{-1mm}
\begin{equation} \label{poly}
{\bf w}^- \leftarrow \tau {\bf w} + (1-\tau){\bf w}^-\vspace{-1mm}
\end{equation}

At each iteration of the training, the parameters of the DQN are updated as
${\bf w}_{k+1} = {\bf w}_{k} + {\bf v}_k$.
The update vector is given as ${\bf v}_k = -\alpha \nabla \mathcal{L}({\bf w}_k)$ where $\nabla \mathcal{L}({\bf w}_k) $ is the gradient of the loss function calculated on a small mini-batch sample drawn at random from the experience replay buffer $\mathcal {D}$. 

\vspace{-2mm}
\section{Nesterov's Accelerated Quasi-Newton Method for Q-learning}
First order gradient based methods have been commonly used in training DQNs due to their simple complexity. However approximated second order quasi-Newton methods have shown to signficantly speed up convergence in non-convex optimization problems. The Nesterov's accelerated quasi-Newton (NAQ) \cite{ninomiya2017novel} and its variants have shown to accelerate convergence compared to the standard quasi-Newton method in various supervised learning frameworks. In this paper, we propose a variant of the Nesterov's accelerated quasi-Newton method and investigate its feasibility in the reinforcement learning framework. The algorithm is shown in Algorithm 1. 

NAQ achieves faster convergence by quadratic approximation of the objective function at ${\bf w}_k+\mu {\bf v}_k$ and by incorporating the Nesterov's accelerated gradient $\nabla \mathcal{L}({\bf w}_k+\mu {\bf v}_k)$ in its Hessian update. The search direction ${\bf {g}}_k=-{\bf {H}}_k \nabla \mathcal{L}({\bf w}_k+\mu {\bf v}_k)$ is computed using the two-loop recursion \cite{Nocedal2006}, where

\vspace{-2mm}
\begin{algorithm}[]

    \centering
\caption{Proposed aSNAQ for DQN}
\begin{algorithmic}[1]
\label{Algo:adaNAQ}
\Require minibatch ${X_k}$, $\mu_{min}, \mu_{max}$, ${k_{max}}$, ${\mathcal{E}_{max}}$, aFIM buffer {\it F} of size ${\it m_F}$ and curvature pair buffer $(S,Y)$ of size ${\it m_L}$, momentum update factor $\phi$, experience replay buffer $\mathcal{D}$
\Ensure ${\bf w}_o $=${\bf w}_k  \in \mathbb{R}^d$, $\mu = \mu_{min}$, ${\bf v}_k $,  ${\bf v}_o $ , ${\bf w}_s $, ${\bf v}_s$,  \& ${t=0}$
\For {episode $\mathcal{E} = 1, 2, ... , \mathcal{E}_{max}$}   
\State Initialize state $s$
\For {step $k = 1, 2, ... , k_{max}$}
\State Take action $a$ based on epsilon greedy strategy 
\State Store transistion $(s, a, r, s', a')$ in  $\mathcal{D}$
\State Sample random minibatch $X_k$ from  $\mathcal{D}$
\State  {Calculate $\nabla \mathcal{L}({\bf w}_k+\mu {\bf v}_k)$} 
\State Determine ${\bf g}_k$ using two loop recursion
\State ${\bf g}_k = {\bf g}_k / ||{\bf g}_k||_2$  
\State ${\bf v}_{k+1}\leftarrow \mu {\bf v}_k +\alpha_k {\bf {g}}_k$
\State ${\bf w}_{k+1}\leftarrow{\bf w}_k +{\bf v}_{k+1}$
\State  {Calculate $\nabla \mathcal{L}({\bf w}_{k+1})$}  and store in $F$
\State {${\bf w}_s = {\bf w}_s + {\bf w}_{k}$} and {${\bf v}_s = {\bf v}_s + {\bf v}_{k}$}

\If {mod(k , L) = 0}
\State Compute avg {${\bf w}_n = {\bf w}_s / L$} and {${\bf v}_n = {\bf v}_s / L$}
\State ${\bf w}_s = 0$ and ${\bf v}_s = 0$
\If {${t>0}$}
\If{$ \mathcal{L}({\bf w}_n) > \eta \mathcal{L}({\bf w}_o)$}
\State {Clear $(S,Y)$ and $F$ buffers}
\State Reset ${\bf w}_k = {\bf w}_o$ and ${\bf v}_k = {\bf v}_o$
\State Update $\mu ={\rm max}(\mu / \phi, \mu_{min})$ 
\State {\bf continue}
\EndIf
\State {$ {\bf s } = {\bf w}_n - {\bf w}_o$}
\State {${\bf y} = \frac{1}{|{\it F}|}(\sum\limits_{i=1}^{|{\it F}|}{\it F_i} \cdot {\bf s})$}
\State Update $\mu ={\rm min}(\mu \cdot \phi, \mu_{max})$ 
\If {${\bf s}^T{\bf y} > \sigma $~${\bf y}^T{\bf y}$}
\State Store curvature pairs ({\bf s},{\bf y}) in $(S,Y)$
\EndIf
\EndIf
\State Update ${\bf w}_o={\bf w}_n$ and ${\bf v}_o={\bf v}_n$
\State $t \leftarrow t+1$
\EndIf
\EndFor
\EndFor
\end{algorithmic}
\end{algorithm}

{\noindent}${\bf H}_k^{(0)}$ is initialized based on the accumulated gradient information given as 
\begin{equation}\label{eqH}
[H_k^{(0)}]_{ii} = \frac{1}{\sqrt{ {\sum_{j=0}^{k} \nabla \mathcal{L}({\bf w}_j)_i^2 } + \epsilon}}.
\end{equation}
aSNAQ uses an accumulated Fisher matrix aFIM for computing the curvature information pair (${\bf s,y}$) for the Hessian computation  as shown in  Eq. (\ref{eqsy}) and Eq. (\ref{eqy})
\begin{equation}\label{eqsy}
 {\bf s} = {\bf w}_{t} - {\bf w}_{t-1}  ,
 \end{equation}
 \begin{equation}\label{eqy}
{\bf y}= \frac{1}{|F|}{ \sum_{i=1}^{|F|}{F}_i \cdot {\bf s}}~,
\end{equation}
where ${\bf w}_t$ is the average aggregated weight, $t$ is the curvature pair update counter, 
 ${F}_i = {\nabla \mathcal{L}({\bf w}_{k+1})}{\nabla \mathcal{L}({\bf w}_{k+1})^T }$ and $|F|$ is the number of $F_i $ entries present in $F$.  The ${\bf y}$ vector is computed without explicitly constructing the ${\nabla \mathcal{L}({\bf w}_{k+1})}{\nabla \mathcal{L}({\bf w}_{k+1})^T }$ matrix by just storing the ${\nabla \mathcal{L}({\bf w}_{k+1})}$ vector. The use of the Fisher Information matrix (aFIM) gives a better estimate of the curvature of the problem. The curvature pair information $({\bf s,y})$ computed based on the average of the weight aggregates and Hessian-vector product reduces the effect of noise and allows for more stable approximations. The curvature pairs are computed every L steps and stored in the $({\bf S},{\bf Y})$ buffer only if sufficiently large. This allows for the updates being made only based on useful curvature information.  Further, the curvature pair information $({\bf s,y})$ and hence the Hessian approximation is updated once in $L$ iterations, thus reducing computational cost.  The size of the $({\bf S},{\bf Y})$ buffer and aFIM buffer $F$ are set to $m_L$ and $m_F$ respectively, thus optimizing storage cost. 

\section{Global Routing}
VLSI physical design requires to compute the best physical layout of millions to billions of circuit components on a tiny silicon surface (${\rm <5cm^2}$). It is carried out in several stages such as partitioning, floor-planning, placement, routing and timing-closure. In the placement stage, the locations of the circuit components, i.e. cells, are determined. Once all cell locations are set, the paths for all the connections of the circuit, i.e. nets, are determined in the routing stage. Global routing involves a large and arbitrary number of nets to be routed, where each net may consist of many pins to be interconnected with wires. In addition, the IC design consideration may impose several constraints such as number of wire crossings (capacity) and routing directions, blockages, etc. The global routing problem can be modelled as a grid maze with multiple start-goal pairs that correspond to the location of the pins to be routed. The objective of the router is to find the optimum connections for all the pins (routing solutions) such that the total wirelength is minimum and no overflow occurs. An overflow is said to occur when the number of wirecrossings exceed the set capacity for a particular edge. For every wire routed, the corresponding capacity decreases. Routing is sequential and hence a common problem is net ordering. Nets routed early can block the routes for later nets due to utilization of the capacity.  The study proposed in \cite{Liao2020} shows potential scope for reinforcement learning based global routing. In this paper we evaluate the efficiency of our proposed algorithm on the reinforcement learning framework for global routing. For each two-pin, the enviroment provides the Q-network with 12 states as input and the output is the estimated action value for all 6 actions as described in \cite{Liao2020}. A reward $R(a,s')$ of +100 is obtained if $s'$ is the target pin, otherwise -1. Further we apply double DQN to enhance the performance.   

\begin{table*}[!htb]

\begin{center}
\caption{Summary of the results on 15 random trials}\label{table}
\label{tab:style}
\fontsize{9}{11}\selectfont
\begin{tabular}{|c|c|ccccc|ccccc|ccccc|}
\hline 
Trial & A* &  &  & Adam &  &  &  &  & RMSprop &  &  &  &  & aSNAQ &  &  \\

Num & WL & WL & diff & $\mathcal{R}_{best}$ & $\mathcal{E}$ & Pins & WL & diff & $\mathcal{R}_{best}$ & $\mathcal{E}$ & Pins & WL & diff & $\mathcal{R}_{best}$ & $\mathcal{E}$ & Pins \\
\hline	1	&	390	&	-	&	-	&	4386	&	465	&	48	&	-	&	-	&	4363	&	490	&	48	&	${\bf 368}$	&	-22	&	4667	&	231	&	50	\\
\hline	2	&	386	&	-	&	-	&	4505	&	399	&	49	&	-	&	-	&	4513	&	483	&	49	&	${\bf 376}$	&	-10	&	4610	&	148	&	50	\\
\hline	3	&	379	&	-	&	-	&	4234	&	478	&	47	&	-	&	-	&	4533	&	401	&	49	&	-	&	-	&	4382	&	344	&	48	\\
\hline	4	&	369	&	348	&	-21	&	4690	&	288	&	50	&	350	&	-19	&	4685	&	492	&	50	&	${\bf 345}$	&	-24	&	4699	&	75	&	50	\\
\hline	5	&	366	&	362	&	-4	&	4679	&	422	&	50	&	${\bf 361}$	&	-5	&	4681	&	430	&	50	&	369	&	+3	&	4656	&	458	&	50	\\
\hline	6	&	352	&	348	&	-4	&	4691	&	437	&	50	&	344	&	-8	&	4697	&	296	&	50	&	${\bf 335}$	&	-17	&	4701	&	157	&	50	\\
\hline	7	&	430	&	-	&	-	&	4053	&	485	&	46	&	-	&	-	&	4322	&	393	&	48	&	-	&	-	&	4324	&	285	&	48	\\
\hline	8	&	398	&	-	&	-	&	4522	&	205	&	49	&	-	&	-	&	4513	&	455	&	49	&	${\bf 377}$	&	-21	&	4663	&	361	&	50	\\
\hline	9	&	369	&	369	&	0	&	4669	&	497	&	50	&	${\bf 347}$	&	-22	&	4687	&	252	&	50	&	348	&	-21	&	4693	&	189	&	50	\\
\hline	10	&	366	&	359	&	-7	&	4674	&	112	&	50	&	375	&	+9	&	4660	&	327	&	50	&	${\bf 357}$	&	-9	&	4683	&	480	&	50	\\
\hline	11	&	379	&	380	&	+1	&	4660	&	252	&	50	&	380	&	+1	&	4658	&	429	&	50	&	-	&	-	&	4523	&	428	&	49	\\
\hline	12	&	351	&	${\bf 346}$	&	-5	&	4692	&	293	&	50	&	351	&	0	&	4689	&	340	&	50	&	348	&	-3	&	4692	&	93	&	50	\\
\hline	13	&	395	&	411	&	+16	&	4616	&	456	&	50	&	397	&	+2	&	4645	&	422	&	50	&	${\bf 394}$	&	-1	&	4640	&	193	&	50	\\
\hline	14	&	340	&	343	&	+3	&	4700	&	409	&	50	&	${\bf 338}$	&	-2	&	4706	&	381	&	50	&	341	&	+1	&	4699	&	49	&	50	\\
\hline	15	&	375	&	374	&	-1	&	4659	&	319	&	50	&	384	&	9	&	4660	&	313	&	50	&	${\bf 371}$	&	-4	&	4668	&	490	&	50	\\

\hline
\end{tabular}
\end{center}\vspace{-5mm}
\end{table*}

\begin{figure}[!b]
\begin{center}\vspace{-5mm}
\includegraphics[width=8cm]{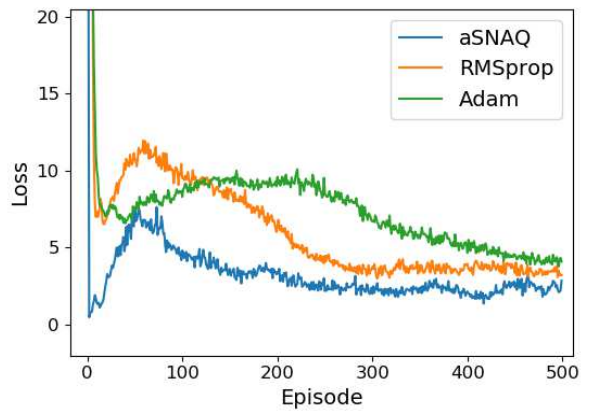}
\end{center}\vspace{-8mm}
\caption{Variation of loss over episodes } 
\end{figure}

\section{Simulation Results}
The performance of the proposed second-order Nesterov's accelerated quasi-Newton method is evaluated on solving global routing. We use the architecture similar to that in \cite{Liao2020}.  In order to further enhance the performance and stability we use double DQN. The neural network structure used is 12--32--64--32--6 with ReLU activation.  A batch size of 32 is chosen. In this paper we consider a two-layer 8x8 grid with a total of 50 nets and maximum two pins in each net. The default capacity is set to 5 and the number of blockages is set to 3. A total of 15 benchmarks were generated using the open-sourced problem set generator \cite{Liao2020}.  The maximum number of episodes was set to 500. An episode constitutes a single pass over the entire set of pin pairs over all nets. The maximum steps is set to 50. The discount factor $\gamma$ is set to 0.9. We evaluate the performance of the proposed aSNAQ method in comparison with Adam and RMSprop. All hyperparameters are set to their default values. The performance metrics include the total wirelength and overflow. For all successful solutions i.e all nets routed within the maximum number of episodes, zero overflow was obtained and the corresponding total wirelength was calculated using the ISPD'08 contest evaluator. The A* search solution is set as the baseline for comparison of the performance metrics. A summary of the results of the 15 benchmarks are shown in Table \ref{table}. The table shows the total wirelength (WL) if all pins were successfully routed. The {\it diff} column shows the amount of wirelength reduction obtained in comparison to the baseline (A* solution) wirelength. $\mathcal{R}_{best}$ indicates the best cummulative reward obtained and $\mathcal{E}$ is the corresponding episode and maximum number of pins successfully routed. 
From the table, it can be observed that for 12 out 15 cases aSNAQ was successful in routing all the pins within 500 episodes  while Adam and RMSprop were successful in only 10 out of 15. Furthermore, in most of the cases the routing solution obtained by aSNAQ had significant wirelength reduction compared to the baseline and in fewer episodes compared to Adam and RMSprop. Fig. 1 shows the average loss over 500 episodes for one of the benchmarks. It can be noted that aSNAQ has the least average loss compared to Adam and RMSprop, thus indicating that aSNAQ is effective in training the deep Q-Network.

\section{Conclusion}
First order gradient based methods are popular in training deep neural networks. Incorporating second order curvature information such as the QN and NAQ methods have shown to be efficient in supervised models. This paper shows the feasibility and efficiency of the proposed stochastic Nesterov's accelerated quasi-Newton (aSNAQ) method in deep reinforcement learning applications as well. Further we apply the proposed algorithm in a deep reinforcement learning framework for global routing. To further enhance the performance, double DQN was used. The results indicate that the DQNs trained using aSNAQ had better routing solutions compared to those trained with Adam and RMSprop and with fewer training episodes.  In future works, further analysis of the proposed algorithm on larger netlists, nets with multipins and study on application to other problems will be studied.

\section*{Acknowledgments}
The authors thank H. Liao et. al. \cite{Liao2020} for the publicly available global routing problem set generator.


\begin{thebibliography}{9}
\bibitem{Sutton1998}
R. S. Sutton and A. G. Barto,
\newblock ``Reinforcement Learning: An Introduction,''
\newblock MIT Press, Cambridge, MA, 1st edition, 1998

\bibitem{kingma2014adam}
D. P. Kingma, and J. Ba,
\newblock ``Adam: A method for stochastic optimization,''
\newblock {\it arXiv preprint} arXiv:1412.6980, December 2014.

\bibitem{RMSprop2012}
T. Tieleman and G. Hinton, 
\newblock ``Lecture 6.5 - RMSProp,'' 
 \newblock COURSERA Neural Networks for Machine Learning. Technical report, 2012.

\bibitem{Nocedal2006}
J. Nocedal and S.J. Wright, 
\newblock ``{\it Numerical Optimization Second Edition}'', \newblock Springer, 2006.

\bibitem{ninomiya2017novel}
H. Ninomiya,
\newblock ``A novel quasi-newton-based optimization for neural network training
  incorporating nesterov's accelerated gradient,''
\newblock {\it NOLTA Journal,} IEICE
  vol. 8, no.4, pp. 289--301, October 2017.

\bibitem{aSNAQ2019}
S. Indrapriyadarsini, S. Mahboubi, H. Ninomiya, and H. Asai, 
\newblock An adaptive stochastic nesterov accelerated quasi-Newton method for
  training RNNs.
\newblock {\it Proc. NOLTA'19}, pp. 208--211, December 2019

\bibitem{Liao2020}
 H. Liao, W. Zhang, X. Dong, B. Poczos, K. Shimada, and L. Burak Kara,  
 \newblock  ``A Deep Reinforcement Learning Approach for Global Routing,'' 
 \newblock {\it Journal of Mechanical Design,} vol. 142 no. 6, June 2020.

\bibitem{DoubleDQN2016}
H. Van Hasselt, A. Guez and D. Silver,
\newblock ``Deep Reinforcement Learning with Double Q-learning,''
\newblock {\it $30^{th}$ AAAI Conf. on Artificial Intelligence,} March 2016


\end{thebibliography}
\end{document}